\documentclass[lettersize,journal]{IEEEtran}
\usepackage{amsmath,amsfonts}
\usepackage{array}
\usepackage[caption=false,font=normalsize,labelfont=sf,textfont=sf]{subfig}
\usepackage{textcomp}
\usepackage{stfloats}
\usepackage{url}
\usepackage{verbatim}
\usepackage{graphicx}
\usepackage{cite}
\usepackage{booktabs} % For improved table formatting
\usepackage{multirow} % For multirow cells in tables
\usepackage{algpseudocode} % For algorithm pseudocode
\usepackage[table]{xcolor}
\usepackage[backref]{hyperref} 
\usepackage{authblk}
\begin{document}

\title{From Efficient Multimodal Models to World Models: A Survey}

\author[1]{Xinji Mai}
\author[1]{Zeng Tao}
\author[1]{Junxiong Lin}
\author[1]{Haoran Wang}
\author[1]{Yang Chang}
\author[1]{Yanlan Kang}
\author[1,*]{Yan Wang}
\author[1,2,3,*]{Wenqiang Zhang}

\affil[1]{Shanghai Engineering Research Center of AI and Robotics, Academy for Engineering and Technology, Fudan University, Shanghai, China}
\affil[2]{Engineering Research Center of AI and Robotics, Ministry of Education, Academy for Engineering and Technology, Fudan University, Shanghai, China}
\affil[3]{Shanghai Key Lab of Intelligent Information Processing, School of Computer Science, Fudan University, Shanghai, China}
\affil[*]{Corresponding author:wqzhang@fudan.edu.cn(Wenqiang Zhang), yanwang19@fudan.edu.cn(Yan Wang)}
% % Corresponding author text

\maketitle

\begin{abstract}
Multimodal Large Models (MLMs) are becoming a significant research focus, combining powerful large language models with multimodal learning to perform complex tasks across different data modalities. This review explores the latest developments and challenges in MLMs, emphasizing their potential in achieving artificial general intelligence and as a pathway to world models. We provide an overview of key techniques such as Multimodal Chain of Thought (M-COT), Multimodal Instruction Tuning (M-IT), and Multimodal In-Context Learning (M-ICL). Additionally, we discuss both the fundamental and specific technologies of multimodal models, highlighting their applications, input/output modalities, and design characteristics. Despite significant advancements, the development of a unified multimodal model remains elusive. We discuss the integration of 3D generation and embodied intelligence to enhance world simulation capabilities and propose incorporating external rule systems for improved reasoning and decision-making. Finally, we outline future research directions to address these challenges and advance the field.
\end{abstract}


\section{The Development Status of Multimodal Models and World Models}
\subsection{World models}

World models are currently one of the hottest research directions in the AI field. From OpenAI to Meta, major AI companies are striving to develop world models. The concept of world models can be traced back to the fields of reinforcement learning and robotic control. Traditionally, reinforcement learning algorithms rely on agents learning through trial and error in real environments, which is not only costly but sometimes infeasible. To overcome these limitations, researchers began exploring methods for simulation and learning within internal environments. Jurgen et al.\cite{ha2018recurrent} have described a method for quickly training through unsupervised environments using generative recurrent neural networks (RNN) to compress spatiotemporal representations and simulate common reinforcement learning environments. Jurgen et al. referred to this as a world model. In AI research, the proposal of world models aims to distinguish this direction from another research focus: agents.

\begin{figure}[htbp]
    \centering
    \includegraphics[width=0.5\textwidth]{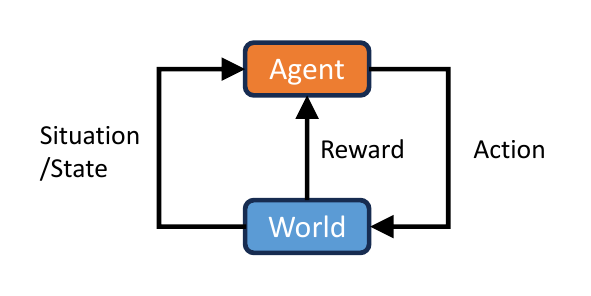}
    \caption{The agent inputs actions into the world simulator, which changes states and outputs feedback. This interaction loop illustrates the process where the agent perceives the current situation, makes decisions, and acts. The world simulator updates its state based on these actions and provides feedback to the agent, essential for learning and adaptation.}
    \label{fig:world_model}
\end{figure}

World models gained widespread attention thanks to Yann LeCun's work\cite{lecun2022path}. Yann LeCun mentioned that human or animal brains seem to run a simulation of the world, which he called a world model. This model helps humans and animals make predictions about their surroundings. LeCun provided an example: a baby learns basic knowledge by observing the world in the first few months after birth, such as understanding gravity when seeing an object fall. This ability to predict what will happen next comes from common sense, which LeCun believes is the essence of intelligence. The Sora model is a model developed by OpenAI for generating video. It utilizes multimodal learning techniques to generate realistic video content by combining text and image data. In recent studies, OpenAI defined Sora in their report as a world simulator capable of generating videos and considered Sora's technology a promising approach to building a general world model. We will introduce the differences between the two main routes currently being explored: the multimodal large models developed by Meta under Yann LeCun's guidance and OpenAI's GPT series.

In summary, we can clearly define a world model as shown in Figure \ref{fig:world_model}. A world model refers to a model that can predict and simulate environmental state changes by learning from various data in the environment. Unlike conventional data testing scenarios where data does not change, a world model's data can change independently, even generating data not in the test dataset. The core function of a world model lies in counterfactual reasoning, which allows it to infer the outcomes of decisions not encountered before. AI researchers' pursuit of world models aims to achieve this counterfactual reasoning, a natural ability of humans that current AI lacks.

\subsection{Routes to World Models}

Currently, there are two main routes to developing world models: autoregressive methods and JEPA (Joint Embedding Predictive Architecture) methods. Autoregressive models hold a significant place in the generative field, with notable representatives including the GPT series and Sora. These models, based on Transformer architecture\cite{vaswani2017attention}, generate data step-by-step, with each output depending on the previous hidden state. This incremental generation allows the model to capture contextual information, producing coherent and logical sequences. The autoregressive model possesses a robust contextual understanding capability and is facile to train, thus establishing itself as the predominant approach in the field of world modeling. By capitalizing on previously generated content during generation, autoregressive models demonstrate adeptness in comprehending and maintaining contextual consistency, thereby yielding more coherent and meaningful output. The training process for autoregressive models is relatively straightforward, involving step-by-step prediction and optimization based on known sequential data, which contributes to their commendable performance when trained on large-scale datasets. While autoregressive models excel in natural language processing tasks, generating high-quality text paragraphs through pre-training and fine-tuning, critics argue that these models lack real-world common sense, obscured by vast amounts of information. For instance, a baby learns how the world works and can predict outcomes with little practice compared to the extensive training data required for large language models.

In response, Meta proposed the JEPA framework. The core idea of JEPA is hierarchical planning, a method of decision-making and control that is especially suited to handling complex tasks and large-scale problems. This approach involves breaking down problems into multiple levels, each addressing sub-tasks at different levels of abstraction, simplifying the overall problem-solving process. LeCun illustrated this with an example: to travel from New York to Beijing, one must first get to the airport, then take a flight to Beijing, with the overall cost function representing the distance from New York to Beijing. Solving this involves decomposing the task into millisecond-level control, finding the action sequence that minimizes the predicted cost. He believes all complex tasks can be accomplished through such hierarchical methods, with hierarchical planning being the most significant challenge.

JEPA models extract abstract representations of the world state through a series of encoders and use different levels of world model predictors to forecast various states at different time scales. Inspired by the human brain's ability to understand and react to the environment in a hierarchical manner, JEPA uses a layered architecture to break down complex tasks into multiple levels, each handling sub-tasks at different abstraction levels. This approach enables JEPA to efficiently capture and predict changes in complex dynamic systems, improving the model's handling of long-time-span and multi-scale data. Its unique hierarchical prediction mechanism not only enhances understanding and prediction accuracy of environmental states but also increases adaptability and robustness in dealing with large-scale, diverse data, showcasing significant advantages in many practical applications.

In summary, we can summarize the route to the world model into two, rules and data drivers.

\subsection{Multimodal Models}

Regardless of the route to world models, multimodal models are an indispensable part. Multimodal models refer to machine learning models capable of processing and understanding data from different modalities, such as images, text, audio, and video \cite{yin2023survey,he2024llms}. Human interaction with the real world involves multiple modalities of information, including language, vision, and audio. Therefore, world models must handle and understand multiple forms of data, meaning they must have multimodal understanding capabilities. Additionally, world models simulate dynamic environmental changes to make predictions and decisions, requiring robust multimodal generation capabilities\cite{jin2024efficient}. To put it simply, the world is multimodal, and the world simulator must be able to accept and generate multimodal information. In essence, world models are general-purpose models (General-Purpose Models).

The research on multimodal models can be broadly categorized into several technical approaches: alignment, fusion, self-supervision, and noise addition. Alignment-based methods map data from different modalities to a common feature space for unified processing. Fusion methods integrate multimodal data at different model layers to fully utilize information from each modality. Self-supervised techniques pre-train models on unlabeled data, enhancing performance across various tasks. Noise addition enhances model robustness and generalization by introducing noise into the data.

Combining these techniques allows multimodal models to demonstrate strong capabilities in handling complex real-world data. They can understand and generate multimodal data, simulate and predict environmental changes, and aid agents in making more precise and effective decisions. Thus, multimodal models play a crucial role in developing world models, marking a key step towards general artificial intelligence (General AI). The following sections will detail the technical routes of multimodal models.

\subsection{Structure of This Paper}

In Section 2, we introduce the fundamental technologies of basic architectures. In Section 3, we will introduce the optimization technology of model architecture. Section 4 covers specific technologies of multimodal models. Section 5 compares and contrasts different routes of multimodal models. Finally, Section 6 outlines the potential development paths for multimodal models towards becoming world models.

\section{Basic Techniques of Multimodal Models}
In this chapter, we will introduce the basic techniques of multimodal models commonly used in two routes, rule-driven and data-driven, from the underlying architecture to the block architecture.
\subsection{Transformers and Their Challengers}
The Transformer architecture is currently one of the most popular deep learning model architectures, especially for natural language processing (NLP) and computer vision (CV) tasks. The Transformer is a deep learning model designed for handling sequential data, employing an attention mechanism to model long-range dependencies within sequences \cite{yin2023survey}. Unlike traditional Recurrent Neural Networks (RNNs), the Transformer processes sequences without relying on their order, utilizing self-attention to simultaneously focus on all positions in the sequence, thereby greatly enhancing parallel computation efficiency. The Transformer consists of an encoder and a decoder, where the encoder maps the input sequence into a continuous representation space, and the decoder generates the output sequence based on this representation. Each layer of the encoder and decoder includes a multi-head self-attention mechanism and a feed-forward neural network, stabilized by residual connections and layer normalization. Due to its efficient parallel computation and powerful representation learning capabilities, the Transformer has achieved remarkable success in natural language processing and other tasks requiring sequence data processing.
However, the self-attention mechanism of Transformers has high computational complexity when processing long sequences, limiting its efficiency in some applications. To address this issue, researchers have proposed various methods to challenge the Transformer architecture.
Linear Attention simplifies the computation of self-attention, reducing its time and space complexity from \(O(N^2)\) to \(O(N)\). Key models include Performer\cite{choromanski2020rethinking}, Linformer\cite{wang2020linformer}, and Linear Transformers\cite{katharopoulos2020transformers}. These models can efficiently handle long-sequence data, reducing computational resource consumption. Additionally, Grouped-query Attention and Multi-query Attention are important attention mechanism variants. The former balances multi-head and multi-query attention by sharing a set of keys and values among groups of query heads, while the latter simplifies computation by sharing the same key and value for all query heads, thus improving efficiency. Grouped Query Attention and Multi-query Attention excel in reducing the size of key-value pairs during inference, significantly improving throughput. Multi-Query Attention shares key-value pairs among multiple heads, achieving a 30-40\% reduction in throughput. Grouped Query Attention groups queries, sharing key-value pairs within groups, achieving results comparable to Multi-Query Attention in both efficiency and performance. MQA and GQA are used in the well-known open-source large language models Llama-2 and Llama-3\cite{touvron2023llama, touvron2023llama2, meta2024introducing}.

At the block level, optimization methods include Compact Architecture, which reduces layers and parameters for a compact model structure, lowering computational costs; Pruning\cite{liu2018rethinking}, which reduces redundant parameters through pruning techniques, enhancing computational efficiency; Knowledge Distillation\cite{gou2021knowledge}, which extracts knowledge from a large teacher model and applies it to a smaller student model, significantly reducing model complexity and computational resource requirements; and Quantization\cite{gray1998quantization}, which converts model parameters from high-precision floating points to lower-precision formats, further reducing computational and storage costs. These optimization methods collectively aim to enhance the efficiency and performance of Transformers, enabling them to process and integrate data from different modalities more efficiently in multimodal tasks.

Additionally, there are architectural approaches challenging the Transformer model, such as Gated Convolution\cite{dauphin2017language} or Gated MLP\cite{rajagopal2021convolutional}, Recurrent Models\cite{mnih2014recurrent}, State Space Models (SSMs)\cite{kalchbrenner2013recurrent,yazici2020orderless}, H3\cite{fu2022hungry}, RWKV\cite{peng2023rwkv}, Mega\cite{ma2022mega}, Yan, JEPA\cite{assran2023self}, and the notable Mamba\cite{gu2023mamba} and Mamba 2\cite{dao2024transformers}. Gated Convolution introduces gating mechanisms in convolutional neural networks (CNNs), enhancing the model's ability to capture local and long-range dependencies while reducing computational load. Recurrent models like LSTM and GRU capture temporal dependencies in sequences through their recursive structures\cite{yu2019review,dey2017gate}, overcoming the vanishing gradient problem in traditional RNNs. State Space Models explicitly model the relationship between system states and observations, providing a flexible framework for handling time-series data, including State Space Models, H3, Mamba and Mamba 2.

These approaches not only offer new theoretical insights but also demonstrate their advantages in practice. This chapter will provide detailed introductions to these basic techniques in general architectures, focusing on the most representative and mainstream approaches.

\subsection{Optimization Techniques for Attention Mechanisms}
Multi-head attention mechanisms are a core component of Transformers, capturing dependencies between different positions in the input sequence through parallel computation of multiple attention heads. However, the standard multi-head attention mechanism has high computational complexity, prompting researchers to propose various variants to optimize its performance.

\begin{figure}[h]
    \centering
    \includegraphics[width=0.5\textwidth]{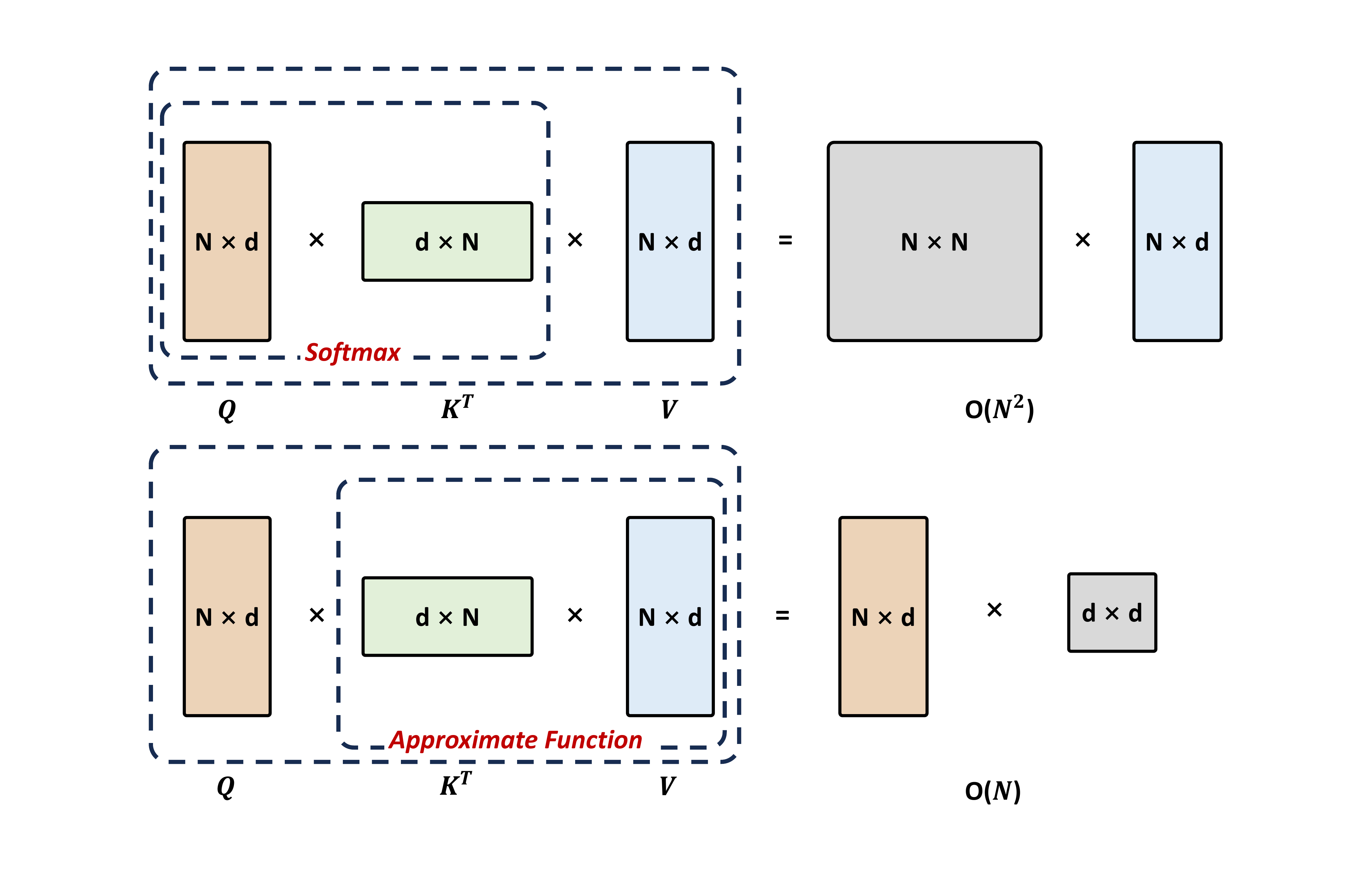}
    \caption{Core principle of simplifying calculations in linear attention mechanisms. The traditional self-attention mechanism with \(O(N^2)\) complexity (top) is replaced by a more efficient linear attention approach with \(O(N)\) complexity (bottom). This is achieved by removing the softmax operation and approximating the function, allowing attention to be computed as a series of matrix multiplications that scale linearly with input size.}
    \label{fig:linear_attention}
\end{figure}

The standard Transformer model faces efficiency bottlenecks when processing long sequences, with the time and space complexity of its self-attention mechanism being quadratic in sequence length \(O(n^2)\). This issue arises from the Softmax operation in the attention mechanism. As shown in Figure \ref{fig:linear_attention}, without Softmax, attention computation simplifies to three matrix multiplications, which are associative, allowing the calculation of \(K^T V\) first, followed by left-multiplying by \(Q\). This reduces complexity from \(O(n^2)\) to linear \(O(n)\), the core idea of Linear Attention.

\begin{equation}
\text{Attention}(Q, K, V) = \text{softmax}\left(\frac{QK^T}{\sqrt{d_k}}\right)V
\end{equation}

Removing Softmax, the attention computation's complexity can drop to linear \(O(n)\). Scaled-Dot Attention essentially weights \(V\) with \(QK^T\). Therefore, a generalized definition of attention can be proposed:

\begin{equation}
\text{Attention}(Q, K, V) = f(QK^T)V
\end{equation}

Here, \(f\) is a general function, approximating the Softmax operation. To fit Softmax, \(f\) must ensure non-negativity \(f \geq 0\). This generalized attention form is known as Non-Local Networks in computer vision.

If elements of \(Q\) and \(K\) are non-negative, their dot product is naturally non-negative. This suggests introducing kernel functions. By adding a non-negative activation function \(\phi\) to \(Q\) and \(K\):

\begin{equation}
\text{Attention}(Q, K, V) = \phi(Q)\phi(K)^T V
\end{equation}

Where \(\phi\) is a non-negative activation function, such as \(\text{ReLU}(x)\). This method, termed the kernel method, is discussed by A Katharopoulos et al\cite{katharopoulos2020transformers}. Performers estimate conventional (softmax) full-rank attention Transformers using linear space and time complexity while retaining provable accuracy, without relying on sparsity or low-rank priors.

Another approach utilizes Softmax's properties. In Efficient Attention\cite{shen2021efficient}, \(Q\) normalized along dimension \(d\) and \(K\) along dimension \(n\) naturally satisfy normalization conditions. Thus, Softmax is applied separately to \(Q\) and \(K\):

\begin{equation}
\text{Attention}(Q, K, V) = \text{softmax}(Q) \cdot \text{softmax}(K)^T V
\end{equation}

This form is a special case of the generalized attention definition. Additionally, sparse attention methods\cite{roy2021efficient}, such as Sparse Attention by OpenAI\cite{child2019generating}, reduce computation by retaining values in local regions only, forcing most attention values to zero. After special design, the non-zero elements of the attention matrix are \(O(n)\), achieving linear-level attention.

Reformer\cite{kitaev2020reformer}, another notable improvement, reduces attention complexity to \(O(n \log n)\) by using locality-sensitive hashing (LSH) to find the largest attention values and compute only those, achieving sparse attention. Moreover, Reformer redesigns the backward propagation process by constructing reversible feed-forward networks (FFN), reducing memory usage. Despite solving sparse attention's first drawback, Reformer remains complex, especially LSH-based attention and reversible network backward propagation.

Performers\cite{choromanski2020rethinking} adopt a novel fast attention method, Fast Attention Via Positive Orthogonal Random features (FAVOR+):

\begin{equation}
\operatorname{Att}_{\leftrightarrow}(\mathbf{Q}, \mathbf{K}, \mathbf{V}) = \mathbf{D}^{-1} \mathbf{A V},
\end{equation}
\begin{equation}
\quad \mathbf{A} = \exp\left(\frac{\mathbf{Q} \mathbf{K}^{\top}}{\sqrt{d}}\right),
\end{equation}
\begin{equation}
\quad \mathbf{D} = \operatorname{diag}\left(\mathbf{A} 1_{L}\right)
\end{equation}

Equivalent to the above attention, the scaling factor \(\sqrt{d_k}\) simplifies \(A\). Fast attention (FA) maps \(Q\) and \(K\) through a \(\phi\) function to \(Q'\) and \(K'\), approximating \(A\) as their product:

\begin{equation}
A = \exp(QK^T) \approx \phi(Q)\phi(K)^T = Q'K'
\end{equation}

\(\phi\) maps matrix row vectors. FAVOR+ simulates beyond softmax other kernelizable attention mechanisms effectively. This capability is crucial for accurately comparing softmax with other kernels on large-scale tasks, aiding in finding optimal attention kernels. Performers are fully compatible with conventional Transformers, offering strong theoretical guarantees: unbiased or nearly unbiased attention matrix estimation, unified convergence, and low estimation variance.

Linformer\cite{wang2020linformer} improves self-attention with low-rank matrix approximation, reducing complexity to linear \(O(n)\). Linformer retains the original Scaled-Dot Attention form but projects \(Q\) and \(K\) to low-dimensional space using \(n \times k\) matrices before attention, reducing computation. While Linformer excels in some tasks, its performance on long-sequence tasks remains to be verified. Moreover, Linformer faces challenges in autoregressive generation tasks due to the projection process combining entire sequence information, complicating causal masking.

\begin{figure}[h]
    \centering
    \includegraphics[width=0.45\textwidth]{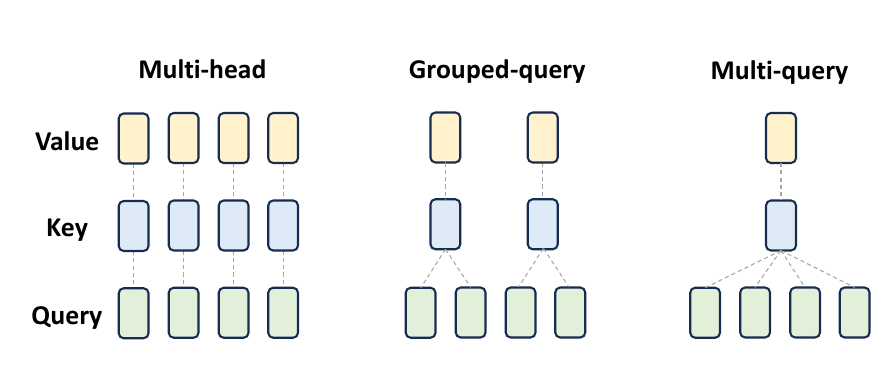}
    \caption{Illustration of Multi-Head Attention (MHA), Multi-Query Attention (MQA), and Group-Query Attention (GQA). This figure demonstrates the structural differences between these attention mechanisms. Multi-Head Attention uses multiple sets of queries, keys, and values to capture different aspects of the input. Grouped-Query Attention shares keys and values among groups of queries to balance computational efficiency and model expressiveness. Multi-Query Attention simplifies the model by sharing the keys and values across all attention heads, reducing the computational load while maintaining performance.}
    \label{fig:attention_variants}
\end{figure}

Multi-Query Attention and Group-Query Attention are notable variants. These methods optimize the attention computation process, reducing computational complexity and memory consumption while maintaining or enhancing model performance. Multi-Query Attention (MQA) shares keys and values among all attention heads, computing independent queries for each head, thus lowering complexity and memory usage. In MQA, all attention heads share the same key and value, differing only in queries:

\begin{equation}
Q_i = QW_i^Q, \quad K = KW^K, \quad V = VW^V
\end{equation}
\begin{equation}
\text{Attention}_i(Q_i, K, V) = \mathrm{softmax}\left(\frac{Q_i K^T}{\sqrt{d_k}}\right) V
\end{equation}

Here, \(Q_i\) is the query for the \(i\)th attention head, \(K\) and \(V\) are shared keys and values. This significantly reduces the number of matrices to compute and store, lowering computational and memory complexity.

As shown in Figure \ref{fig:attention_variants}, Group-Query Attention differs from MQA by grouping attention heads, with each group sharing the same keys and values, while heads within each group have independent queries. This method reduces complexity while maintaining flexibility:

\begin{equation}
Q_{i,j} = QW_{i,j}^Q, \quad K_i = KW_i^K, \quad V_i = VW_i^V
\end{equation}
\begin{equation}
\text{Attention}_{i,j}(Q_{i,j}, K_i, V_i) = \mathrm{softmax}\left(\frac{Q_{i,j} K_i^T}{\sqrt{d_k}}\right) V_i
\end{equation}

Here, \(Q_{i,j}\) is the query for the \(j\)th attention head in the \(i\)th group, \(K_i\) and \(V_i\) are the shared keys and values for the \(i\)th group. This group-sharing approach provides more flexibility and expressiveness while reducing computational and memory complexity.

Additionally, down-sampling techniques such as pooling or using strided 1D convolutions reduce sequence length. IBM's PoWER-BERT\cite{goyal2020power} and Google's Funnel-Transformer\cite{dai2020funnel} improve model efficiency through gradual down-sampling. Although these techniques significantly reduce complexity, they can impact the model's generative capability.

Overall, modifying attention forms or structures effectively reduces computational complexity while retaining or enhancing performance. These variants offer more efficient solutions for handling long-sequence data and provide new ideas and methods for multimodal task research.

\section{Optimization Techniques for Model Architectures}

In the research of multimodal methods, addressing issues such as excessive parameters in Transformer models, researchers have proposed various optimizations and improvements to enhance model efficiency, reduce computational complexity, and improve performance. This section provides a detailed introduction to these improvements in model architectures.

\subsection{Model Compression}
Model compression aims to reduce the number of parameters and computational load in deep neural networks, enhancing efficiency and reducing storage requirements. Pre-trained deep neural network models often face over-parameterization, where only about 5\% of the parameters are effective. Model compression techniques include frontend and backend compression, aiming to shrink model size without significantly reducing accuracy, thus improving usability and efficiency in practical applications.

Frontend compression methods include knowledge distillation, compact model structure design, and filter pruning. Knowledge distillation transfers knowledge from a complex model to a smaller one, allowing the small model to maintain high computational efficiency while achieving the performance of the complex model. For example, Hinton et al. \cite{hinton2015distilling} proposed knowledge distillation techniques that transfer teacher model knowledge through softened output probability distributions. Compact model structure design improves the convolution method of neural networks (e.g., using depthwise separable convolutions) to reduce computational parameters. MobileNet \cite{howard2019searching} is a successful example in this aspect. Filter pruning removes unimportant weight matrices to reduce model redundancy.

Backend compression methods include low-rank approximation and unrestricted pruning. Low-rank approximation reconstructs large weight matrices with several low-rank matrices, reducing storage and computational resource consumption. For example, Singular Value Decomposition (SVD)\cite{stewart1993early} is widely used for matrix decomposition to achieve compression. Unrestricted pruning includes unstructured pruning and structured pruning. Unstructured pruning removes individual weights.

In addition to these traditional compression methods, Han Song's team proposed AutoML Model Compression (AMC) \cite{he2021automl}, utilizing reinforcement learning to automatically search for model compression strategies, enhancing the efficiency of deploying neural network models on mobile devices. AMC uses reinforcement learning to intelligently balance model size, speed, and accuracy, automatically generating optimal compression strategies more efficiently and effectively than manually crafted heuristic rules.

These model compression techniques improve computational and storage efficiency, allowing deep neural networks to be widely applied in resource-constrained environments.

\subsection{Model Pruning}
Pruning techniques remove redundant parameters and connections in a model to enhance computational efficiency and reduce model size. Pruning techniques can be categorized into unstructured pruning, structured pruning, and hybrid pruning.

Unstructured pruning operates at a fine granularity, removing arbitrary "redundant" parameters in the network. However, this method may result in irregular network structures that are difficult to accelerate effectively. LeCun proposed the Optimal Brain Damage (OBD) algorithm \cite{lecun1989optimal} in the late 1980s, using the second-order derivatives of the loss function to determine parameter importance. Hassibi et al. \cite{hassibi1992second, hassibi1993optimal} extended this with the Optimal Brain Surgeon (OBS) algorithm, not limited by OBD's diagonal assumption, zeroing out less important weights and recalculating others to compensate for activation values, achieving better compression results. Srinivas et al. \cite{srinivas2015data} proposed methods to remove dense connections in fully connected layers without relying on training data, significantly reducing computational complexity. 

Structured pruning removes structural components based on predefined criteria, such as attention heads or layers.X-Pruner \cite{yu2023x} utilizes explainable masks learned end-to-end, measuring each unit's contribution to predicted target classes, and adaptively searches layer-wise thresholds to retain the most informative units while determining pruning rates. 

Hybrid pruning combines unstructured and structured pruning methods, balancing their advantages for better performance optimization. For example, SPViT \cite{kong2022spvit} developed a multi-head token selector based on dynamic attention for adaptive instance-level token selection, introducing a soft pruning technique that merges less important tokens into packet tokens rather than discarding them. ViT-Slim \cite{zheng2022savit} introduced learnable and unified sparsity constraints with predefined factors to represent global importance within various dimensions.

These pruning techniques demonstrate significant effectiveness in various applications. In image classification tasks, structured pruning can significantly reduce convolutional neural networks' computational costs while maintaining high classification accuracy. In natural language processing tasks, unstructured and hybrid pruning effectively reduce Transformer model complexity, enabling inference and training with lower resource consumption. By applying these pruning techniques, models maintain high performance while significantly reducing computational and storage costs, enhancing their usability and efficiency in practical applications.

\subsection{Knowledge Distillation}
Knowledge Distillation (KD) is a model compression technique that transfers knowledge from a complex model (called the teacher model) to a smaller model (called the student model). This allows the student model to maintain high computational efficiency while achieving the performance of the teacher model. Knowledge distillation was first proposed by Buciluǎ et al., who trained compressed models with pseudo-data classifiers to replicate the original classifier's outputs\cite{gou2021knowledge}. KD can be divided into homomorphic KD and heteromorphic KD.

Homomorphic KD means the student and teacher models have similar or identical structures. In this approach, the student model learns by mimicking the teacher model's outputs (e.g., logits, feature layer outputs). Common homomorphic KD methods include logit-level distillation, feature-level distillation, and module-level distillation. For instance, TinyViT\cite{wu2022tinyvit} applies distillation during pre-training, storing logits from a large teacher model on hardware to achieve memory and computational efficiency when transferring knowledge to a smaller student Transformer. DeiT-Tiny\cite{wang2022towards} adopts patch-level distillation, training a small student model to match the pre-trained teacher model's patch structure, then optimizing with decomposed manifold matching loss to reduce computational costs. Module-level methods like m2mKD\cite{lo2024m2mkd} separate the teacher module from a pre-trained unified model, combining student modules with modular models, and using a shared meta-model for composition, enabling student modules to mimic teacher module behavior. Feature-level distillation methods like MiniViT\cite{zhang2022minivit} combine weights from consecutive Transformer blocks for cross-layer weight sharing, introducing transformations to enhance learning.

Heteromorphic KD refers to student and teacher models with different structures. In this approach, the student model learns by mimicking the teacher model's outputs or intermediate features, despite different architectures. Heteromorphic KD enhances the student model's adaptability, enabling it to learn useful information from the teacher model. Heteromorphic KD includes soft label distillation, where the student model trains by mimicking the teacher model's soft label outputs.

KD transfers knowledge from complex models to smaller models, achieving model compression and acceleration. Both homomorphic and heteromorphic KD train by mimicking the teacher model's outputs or features. These methods not only improve student model performance but also reduce computational and storage costs, enabling deep learning models to be widely applied in resource-constrained environments. Studies show that models processed by KD can perform well in resource-constrained environments such as mobile devices and embedded systems, further promoting deep learning technology's wide deployment in practical applications.

\subsection{Quantization Techniques}
Quantization techniques convert model parameters from high-precision floating-point numbers (e.g., 32-bit or 64-bit) to lower-precision formats (e.g., 8-bit or 16-bit), reducing computational and storage costs\cite{liang2021pruning,gholami2022survey}. For example, when training a cat-dog classification model on a laptop, its parameter size might be 64MB. Deploying it on an Arduino Uno using an ATmega328P microcontroller with 8-bit operations, quantizing the model can reduce the weight storage size to 1/8 of the original, with negligible accuracy impact (about 1-3\%). This demonstrates quantization's significant advantages in reducing storage needs and improving computational efficiency.

Weights are trainable parameters in neural networks, adjusted during training to minimize the model's loss function, enabling the model to learn from data. Each layer's weights transform input features to output features through matrix multiplication. Suppose the input vector is \(\mathbf{x}\), weight matrix \(\mathbf{W}\), and bias vector \(\mathbf{b}\), then the neural network layer output can be represented as:

\begin{equation}
\mathbf{z} = \mathbf{W} \mathbf{x} + \mathbf{b}
\end{equation}

Quantization techniques convert model parameters from high-precision floating points to lower-precision formats, effectively reducing computational and storage costs. Quantization methods include post-training quantization, quantization-aware training, and hardware-aware quantization.

Post-Training Quantization (PTQ) quantizes model parameters after training completion. PTQ's main advantage is simplicity and speed, not requiring adjustments to the training process. It typically includes weight and activation quantization. Quantization-Aware Training (QAT) considers quantization impacts during training, simulating quantization effects to help the model adapt to post-quantization performance. QAT typically includes weight and activation quantization. Hardware-Aware Quantization (HAQ) considers specific hardware architecture characteristics during quantization, optimizing model performance on specific hardware. HAQ not only considers scaling factors but also combines hardware features for optimization, such as adjusting scaling factors and quantization ranges to suit hardware characteristics.

The industry widely adopts INT8 quantization, replacing FP32 during inference while training still uses FP32. Many deep learning software such as TensorRT, TensorFlow, PyTorch, and MxNet have enabled or are enabling quantization. Quantization techniques enable deep learning models to be widely applied in resource-constrained environments while maintaining high computational performance and low storage requirements.

\subsection{Synthetic Data Techniques}
Synthetic data techniques generate data similar to real data but without containing real personal information, expanding training datasets to improve model generalization and robustness\cite{bolon2013review,jordon2022synthetic}. In large model training, pure text synthesis is mostly done through other large models, while image synthesis mainly uses generative models. Synthesized data often needs to be validated through statistical methods to ensure conformity with real data distribution.

Statistical methods generate synthetic data by performing statistical analysis and modeling on real data, then using these models to generate synthetic data. For example, using probability distribution functions to simulate real data characteristics and distribution to generate synthetic data. Generative Adversarial Networks (GANs) are deep learning techniques used to generate realistic synthetic data. GANs consist of a generator and a discriminator, where the generator produces synthetic data, and the discriminator distinguishes between real and synthetic data. Through continuous adversarial training, the generator and discriminator compete, ultimately generating high-quality synthetic data. GANs have wide applications in medical imaging, facial recognition, and autonomous driving. Variational Autoencoders (VAEs) are generative models that learn data latent representations to generate synthetic data similar to real data distribution. VAEs are particularly suitable for image generation tasks, performing well in generating high-quality, realistic images. Sequence models generate synthetic data for sequence data (e.g., text, time series) through models such as Markov Chains, Recurrent Neural Networks (RNNs), and Variational Autoencoders (VAEs), modeling sequence features and dependencies to generate synthetic data.

\subsection{Evaluation Techniques for Model Architectures}
Evaluation techniques for model architectures measure and compare the performance of different deep learning models to select the best architecture. Evaluation methods can be divided into manual and automatic evaluations.

Manual evaluation involves experts or users assessing model outputs, suitable for tasks with strong subjectivity, such as the quality of generated text or the realism of generated images. However, manual evaluation is inefficient, costly, and difficult to scale.

Automatic evaluation measures model performance by computing various performance metrics, offering high efficiency and repeatability. Common automatic evaluation platforms and tools include Prompt Flow in Microsoft Azure AI Studio, Weights Biases combined with LangChain, LangSmith\cite{ito2020langsmith} in LangChain, DeepEval\cite{
} in Confidence-ai, and TruEra\cite{ono2024evaluating}. These platforms and tools provide various evaluation methods, such as rule-based and model-based evaluations.

Rule-based evaluation methods use predefined rules and metrics (e.g., accuracy, precision, recall, F1 score, ROC-AUC curve) to assess model performance. For example, datasets like MMLU\cite{wang2024mmlu}, TriviaQA\cite{joshi2017triviaqa}, and HumanEval\cite{chen2021evaluating} are widely used to evaluate language model understanding and generation capabilities. Model-based evaluation methods use pre-trained referee models (e.g., GPT-4, Claude) or adversarial evaluation (e.g., LLM Peer-examination) to assess model performance. These methods comprehensively evaluate model performance on complex tasks and multimodal data.

\subsection{Fine-Tuning Techniques for Model Architectures}
Fine-tuning techniques involve further training pre-trained models on specific task datasets to enhance model performance in that task. Below are common fine-tuning techniques and their recent advancements:

LoRA (Low-Rank Adaptation)\cite{hu2021lora} is a low-rank adaptation technique that adds low-rank matrices to pre-trained models for fine-tuning, reducing computational and storage costs while maintaining performance. QLoRA\cite{dettmers2024qlora} is an improved version that further optimizes the fine-tuning process through quantization techniques. Retrieval-Augmented Generation (RAG)\cite{lewis2020retrieval} combines information retrieval and generative models, enhancing generative model performance by retrieving relevant information from external data sources. The LangChain\cite{topsakal2023creating} library provides various tools allowing large models to access real-time information from sources like Google Search, vector databases, or knowledge graphs, further improving RAG effectiveness. LlamaIndex (GPT Index)\cite{pinheiro2023construction,rogers2017examining} is an integrated data framework designed to enhance large language models (LLMs) by enabling the use of private or custom data. LlamaIndex provides data connectors, indexing and graph-building mechanisms, and advanced retrieval and query interfaces, simplifying data integration and information retrieval processes.

By applying these fine-tuning techniques appropriately, pre-trained model knowledge can be fully utilized, improving performance in new tasks while reducing training time and computational resource consumption.

\subsection{Other Challengers to Model Architectures}

In the field of multimodal large models, the Transformer architecture is widely used for its excellent performance and flexibility. However, as model size and application demands increase, the Transformer architecture faces challenges in computational complexity and memory bottlenecks. To address these challenges, researchers have proposed various optimization strategies and alternative architectures to enhance model efficiency and scalability. Besides Transformers, most other challenger architectures originate from recurrent neural networks (RNNs), including Gated Convolution, Temporal Convolutional Networks (TCN), RWKV, Mamba, and S4, which replace attention with recurrent structures. This approach uses fixed memory to remember previous information, although it can remember a certain length, achieving longer lengths is challenging. Another approach is improving Transformers, such as linear attention improvements mentioned earlier. Representative models include Mega, Yan, and JEPA. We will introduce some representative approaches among them.

The RWKV model\cite{peng2023rwkv} uses linear attention mechanisms, allowing the model to parallelize computations during training and maintain constant computational and memory complexity during inference. The RWKV model consists of stacked residual blocks, each containing time-mixing and channel-mixing sub-blocks, using a recurrent structure to leverage past information. The authors trained RWKV models with sizes ranging from 169 million to 14 billion parameters, making it the largest dense RNN trained to date. Experimental results show that RWKV performs comparably to Transformers of similar size, indicating future work can utilize this architecture to create more efficient models. However, RWKV models have some limitations, such as linear attention potentially limiting performance on tasks requiring long-term dependencies.

The Mega model\cite{ma2022mega} introduces sparse attention mechanisms, zeroing out most elements in the attention matrix and retaining only a few important attention values. This method significantly reduces computational load and memory usage while maintaining predictive performance. Similar to Longformer and Sparse Transformer, the Mega model has unique optimizations in sparse strategies and implementations. By using sparse attention mechanisms, the Mega model greatly reduces computational complexity and memory usage, making it more efficient in handling long-sequence tasks.

JEPA (Joint Embedding Predictive Architecture)\cite{assran2023self} is a novel machine learning model designed to optimize complex tasks and large-scale problem handling through hierarchical decision-making and control methods. The core idea is to decompose problems into multiple layers, each handling subtasks at different abstraction levels, simplifying the overall problem-solving process. The concept and research of JEPA are mainly proposed by Yann LeCun's team at Meta, aiming to overcome the limitations of current large language models (LLMs) in handling complex tasks. A representative method is I-JEPA, a non-generative self-supervised learning method that learns highly semantic image representations by predicting representations of different target blocks in the same image from a single context block. This novel architecture combines the strengths of RNNs and Transformers while reducing their limitations.

Mamba and Mamba 2\cite{gu2023mamba,dao2024transformers} are key directions of improvement. Mamba is an improvement of SSM. State Space Models (SSM) describe dynamic systems and are widely used in control theory, signal processing, and statistical modeling. SSM uses state variables to represent the system's internal state, described by state and output equations. The state equation is:

\begin{equation}
h(t+1) = A h(t) + B x(t) + w(t)
\end{equation}

where \( h(t) \) is the state vector at time \( t \), \( A \) is the state transition matrix, \( x(t) \) is the input vector at time \( t \), \( B \) is the input matrix, and \( w(t) \) is process noise, usually assumed to be zero-mean Gaussian white noise.

The output equation is:

\begin{equation}
y(t) = C h(t) + D x(t) + v(t)
\end{equation}

where \( y(t) \) is the output vector at time \( t \), \( C \) is the output matrix, \( D \) is the direct transmission matrix, and \( v(t) \) is measurement noise, usually assumed to be zero-mean Gaussian white noise.

\begin{figure}[h]
    \centering
    \includegraphics[width=0.5\textwidth]{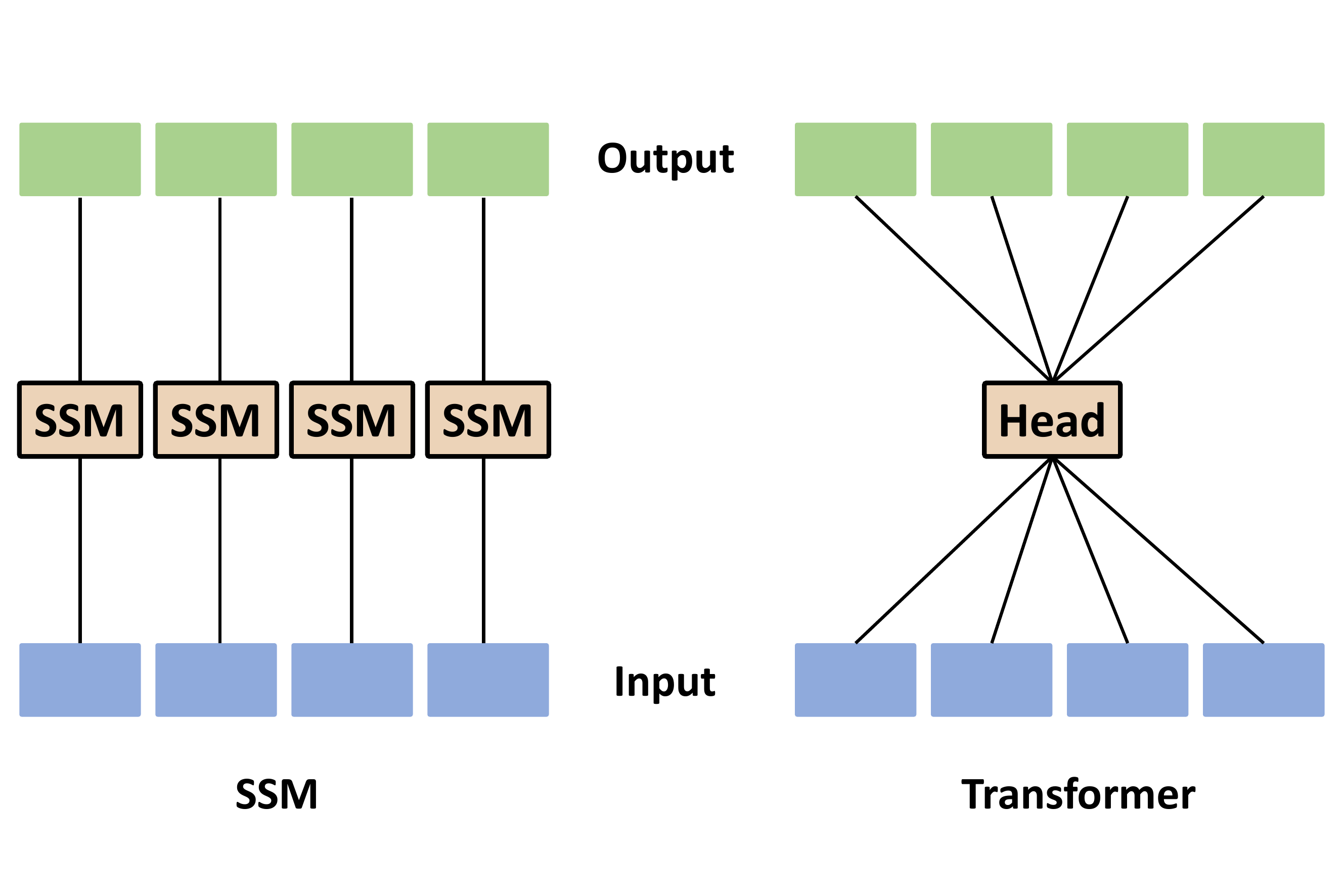}
    \caption{Architectural differences between SSM and Transformer in processing multidimensional inputs. The SSM (Selective State Space Model) processes each input dimension independently, allowing for high parallel computation and linear complexity. In contrast, the Transformer uses a multi-head attention mechanism to capture global dependencies across input dimensions, resulting in higher computational complexity but more comprehensive contextual understanding.}

    \label{fig:ssm_transformer_diff}
\end{figure}

Mamba is a Selective State Space Model (SSSM) based on SSM improvements. Figure \ref{fig:ssm_transformer_diff} shows the architectural differences between SSM and Transformer when handling multidimensional input data. SSM processes each dimension independently, with high parallel computing capabilities and linear computational complexity. In contrast, Transformers capture global dependencies through multi-head attention mechanisms but have higher computational complexity.

\begin{figure*}[h]
    \centering
    \includegraphics[width=7in,height=4in]{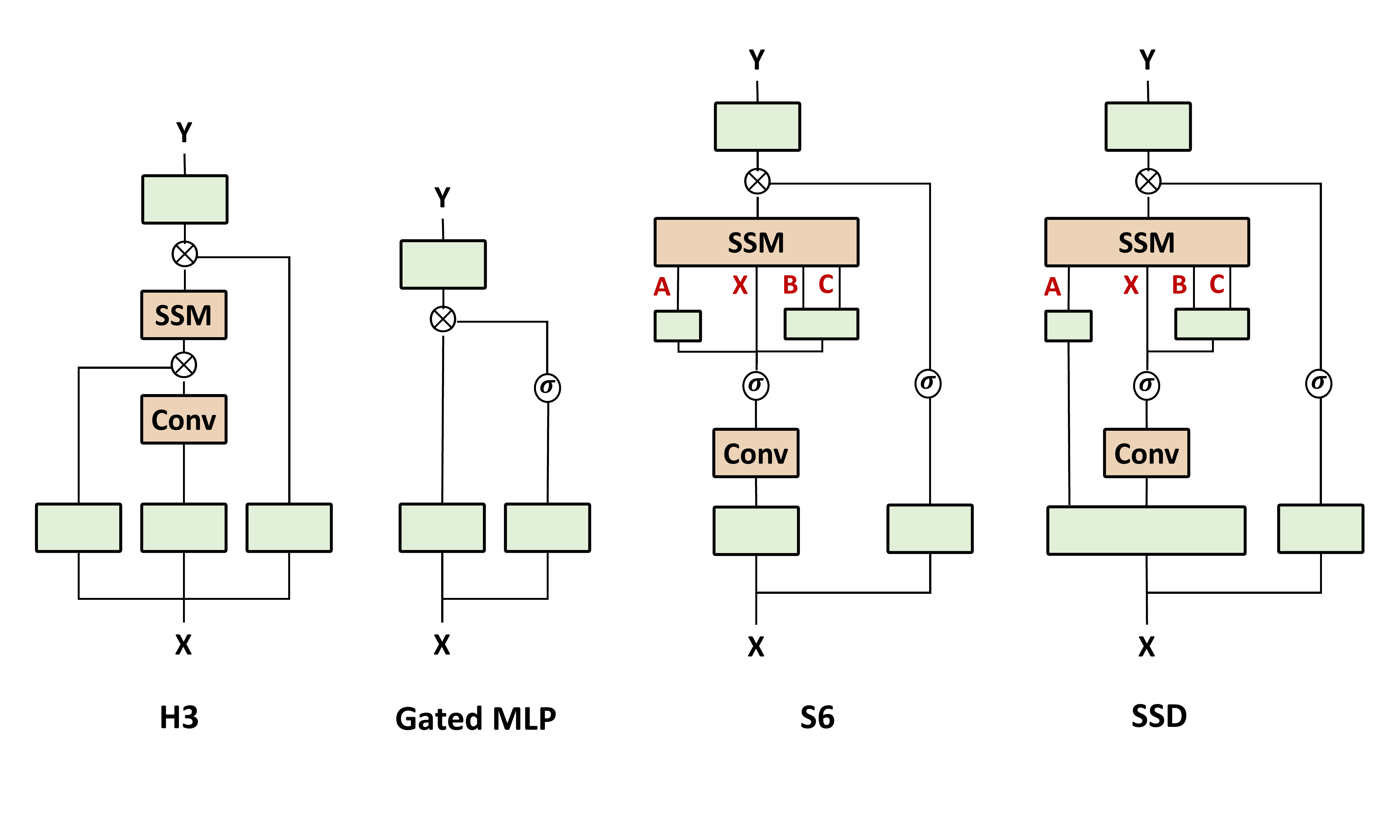}
    \caption{H3 structure compared with SSM. The figure shows different architectures like H3, Gated MLP, S6, and SSD, explaining how each configuration incorporates selective state space models (SSM) and convolutional layers to handle inputs and generate outputs efficiently.}
    \label{fig:h3_block}
\end{figure*}

H3 architecture\cite{fu2022hungry}, a foundational design for homogenized architecture, improves the initial SSM structure, addressing SSM's challenge of remembering long-term data. As shown in Figure \ref{fig:h3_block}, researchers merged the previous SSM architectural design H3 with Gated MLP blocks into one block, selectively processing input information (Selection Mechanism), simplifying the deep sequential model architecture, forming a simple, homogeneous architecture with selective state spaces (Mamba). Like structured SSM, selective SSM is an independent sequence transformation, flexibly integrating into neural networks.

Mamba-2 proposes the State Space Duality (SSD) framework. Based on this, researchers designed the new Mamba-2 architecture, with its core layer being an improved selective SSM. Researchers mixed 4-6 attention layers with Mamba-2 layers, outperforming Transformer++ and pure Mamba-2, indicating attention and SSM are complementary.

One main goal of Mamba-2 is to accelerate SSM using tensor cores. In Mamba-2's SSD structure, parallel projections are realized, breaking through SSM's sequential computation limitations to achieve parallel computation. In actual architecture changes, some SSM parameters being internal activation functions (states) rather than layer input functions limit parallel computation and training speed. In Mamba-2, all SSM parameters are layer input functions, easily applying tensor parallelism to input projections.

Researchers trained a series of Mamba-2 models on the Pile dataset, showing Mamba-2 matches or exceeds Mamba and open-source Transformers in standard downstream evaluations. For instance, a 2.7B parameter Mamba-2 trained on 300 billion tokens in the Pile dataset outperforms the 2.8B parameter Mamba and Pythia, as well as the 6.9B parameter Pythia.

\section{Specific Techniques of Multimodal Models}

\subsection{Multimodal Architecture Techniques}

\begin{figure}[h]
    \centering
    \includegraphics[width=0.5\textwidth]{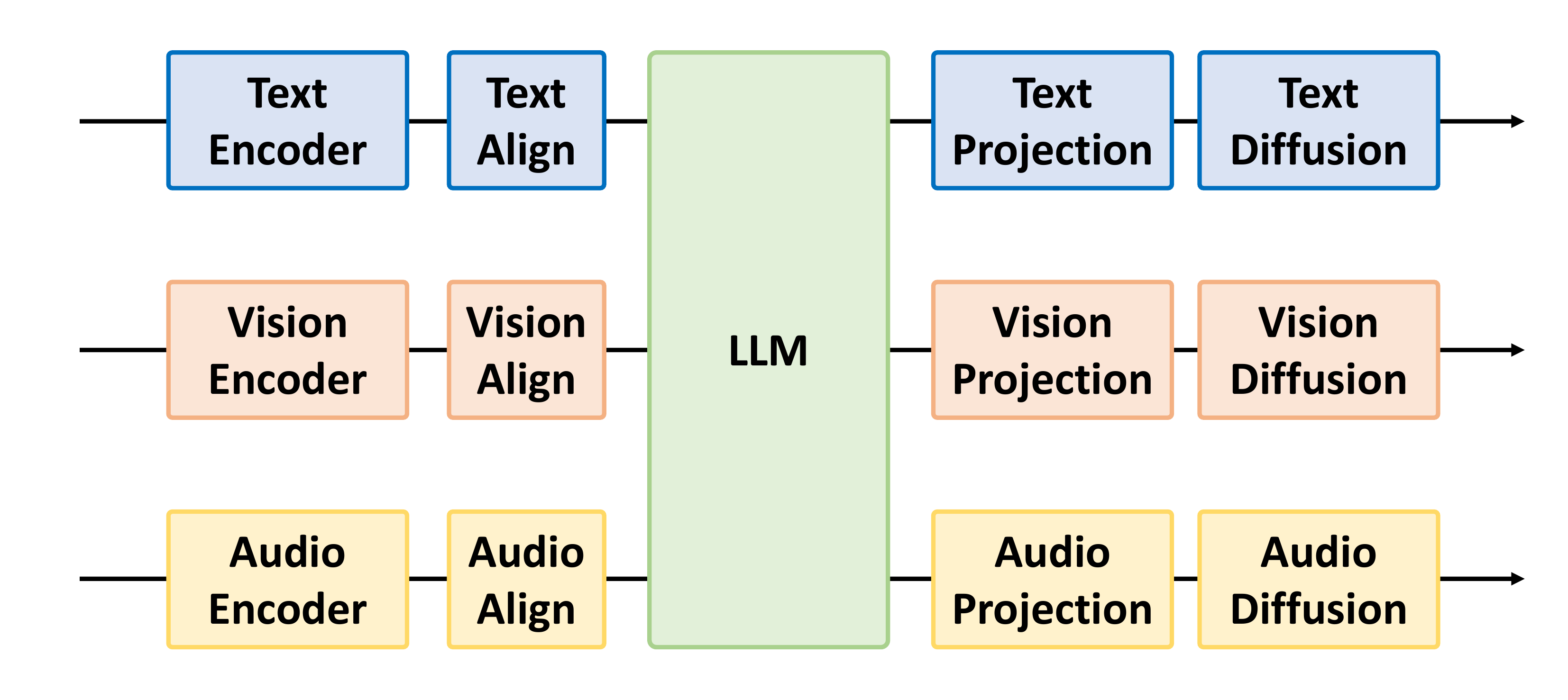}
    \caption{General architecture of a multimodal large model. This figure illustrates how different encoders (Text, Vision, Audio) process their respective inputs and align them before projecting into a shared latent space managed by a Large Language Model (LLM). The model then performs diffusion processes to generate or refine outputs, integrating information across multiple modalities to enhance understanding and generation capabilities.}
    \label{fig:multimodal_architecture}
\end{figure}

\subsubsection{General Multimodal Architectures and Training Strategies}

In the field of multimodal large models (MLM), researchers have proposed various architectural techniques to achieve and optimize the performance and application of multimodal models. Figure \ref{fig:multimodal_architecture} shows a general architecture designed to handle data from text, vision, and audio modalities. In this architecture, each modality's data is first processed through its respective encoder (Text Encoder, Vision Encoder, Audio Encoder) for feature extraction. The features are then normalized and matched through alignment modules (Text Align, Vision Align, Audio Align), followed by projection modules (Text Projection, Vision Projection, Audio Projection) to map the features into a common feature space. Finally, diffusion modules (Text Diffusion, Vision Diffusion, Audio Diffusion) further propagate and adjust the features. The large language model (LLM) integrates these multimodal features to handle and generate complex cross-modal tasks.

This design allows different modalities of data to be fused and processed in a unified feature space, enhancing the understanding and generation capabilities of multimodal data. Specialized modules for encoding, alignment, projection, and diffusion enable the LLM to efficiently process and integrate text, vision, and audio data, thus improving overall model performance and applicability.

End-to-end learning is a crucial training strategy for multimodal large models, where the entire model is optimized as a whole, rather than in stages. Compared to stage-wise training, end-to-end learning eliminates intermediate data processing and model design at each step. However, end-to-end learning for multimodal large models has three major drawbacks.

The two biggest drawbacks are the requirement for large amounts of data and computing power. Direct end-to-end learning necessitates vast multimodal datasets and computational resources. For example, OpenAI used approximately 2.15e25 FLOPS, about 25,000 A100 GPUs, training for 90 to 100 days, with an efficiency (MFU) of about 32\% to 36\% for GPT-4 training, which included about 1.3 trillion tokens. For full multimodal training, these requirements would at least double.

The final drawback is the difficulty in establishing complex relationships. Manually designed modules often inject human prior knowledge, such as encoders, decoders, alignment layers, etc., which can simplify models. For instance, if we aim to detect micro-expressions through video, the model design typically involves keyframe selection, face cropping, facial action unit recognition, combined with micro-expression theory and statistics. An end-to-end model directly establishing connections between images and micro-expressions is evidently challenging and complex.

\begin{figure*}[h]
    \centering
    \includegraphics[width=6.5in]{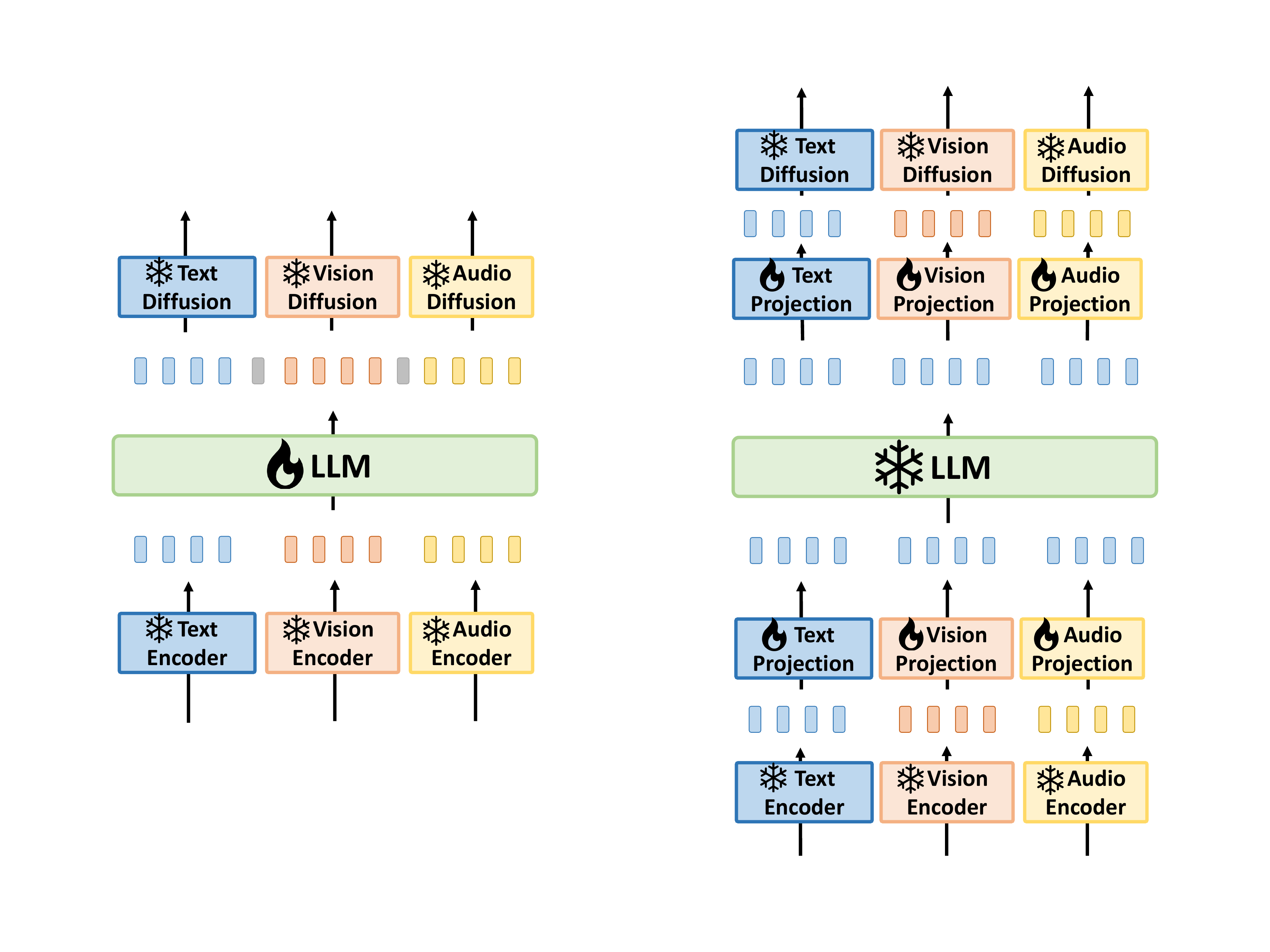}
    \caption{Training strategies for multimodal large models. The left side illustrates the cold start training strategy, where the model is trained from scratch, utilizing text, vision, and audio encoders to process and align data before passing it through the Large Language Model (LLM) and diffusion modules. The right side demonstrates the warm start training strategy, which leverages pre-trained LLMs. Here, pre-trained text, vision, and audio projections are used to refine and enhance the model's performance, allowing for more efficient training by building upon existing knowledge.}

    \label{fig:training_strategies}
\end{figure*}

Given these challenges, most multimodal large models do not entirely use end-to-end training. Figure \ref{fig:training_strategies} shows two training strategies used in large model training. The left side shows the Cold Start Training strategy, where the model trains from scratch. It starts with encoding data from different modalities using text, vision, and audio encoders, followed by feature propagation through diffusion modules (Text Diffusion, Vision Diffusion, Audio Diffusion), then integrates them using a large language model (LLM), and finally projects features through projection modules (Text Projection, Vision Projection, Audio Projection) to generate output. The process emphasizes gradually expanding and adjusting features, ensuring effective integration and processing of multimodal data.

The right side shows the Warm Start Training strategy, where the model starts with some pre-training. The pre-trained LLM directly processes input data through projection modules (Text Projection, Vision Projection, Audio Projection), generates initial features, and refines them through diffusion modules (Text Diffusion, Vision Diffusion, Audio Diffusion). Compared to cold start, warm start leverages existing knowledge from pre-trained models, improving training efficiency and initial performance, suitable for scenarios with relevant domain knowledge or foundational models. This approach enables models to quickly adapt to new tasks and exhibit high performance early in training.

\subsubsection{General Multimodal Encoders}

In terms of vision encoders, consistent with mainstream MLM practices, the pre-trained CLIP model is usually chosen for visual encoding because it effectively aligns the feature spaces of visual and textual inputs. Given the relatively small proportion of visual encoders in MLM parameters, lightweight optimization is less critical compared to language models. By combining multiple visual encoders, a broad range of visual representations can be captured, enhancing model understanding. For example, Cobra\cite{zhao2024cobra} integrates DINOv2 and SigLIP as its visual backbone, combining DINOv2's low-level spatial features with SigLIP's semantic attributes. SPHINX-X\cite{gao2024sphinx} uses two visual encoders, DINOv2 and CLIP-ConvNeXt, pre-trained with different methods and architectures to provide complementary visual knowledge.
Efficient visual encoding models use techniques like token processing to manage high-resolution images without excessive computational burden. High-resolution images are input into lightweight visual encoders, resized, and segmented to generate initial visual tokens. These tokens are compressed by visual token compression modules to reduce computational and storage overhead. Compressed tokens are projected into the language model's feature space by efficient visual-language projectors, aligned with text tokens. A small language model combines and processes these aligned visual features and text tokens, generating language responses. LLaVA-UHD \cite{xu2024llava} introduces an image modular strategy, dividing images into smaller fragments for efficient encoding, reducing computational load while maintaining perceptual ability. Advances in visual encoders also include MAE (Masked Autoencoders)\cite{he2022masked}, a self-supervised learning method that learns image representations by masking and reconstructing parts of input images.

Text encoders are another crucial component of multimodal models, used to process and understand textual data. Transformers are a common text encoding architecture, with self-attention mechanisms efficiently capturing long-range dependencies in text. BERT (Bidirectional Encoder Representations from Transformers) is a pre-trained model based on Transformers, generating high-quality text representations through bidirectional training on large-scale corpora, widely applied in various natural language processing tasks.

In audio encoding, AudioCLIP\cite{guzhov2022audioclip} is an effective choice, generating audio representations by combining audio and text information. AudioCLIP uses an architecture similar to CLIP, aligning audio, text, and image features in the same feature space through contrastive learning. This method enhances audio data representation and improves multimodal model performance in audio-text and audio-image tasks.

\subsubsection{General Multimodal Generative Models}

\begin{figure*}[h]
    \centering
    \includegraphics[width=7in]{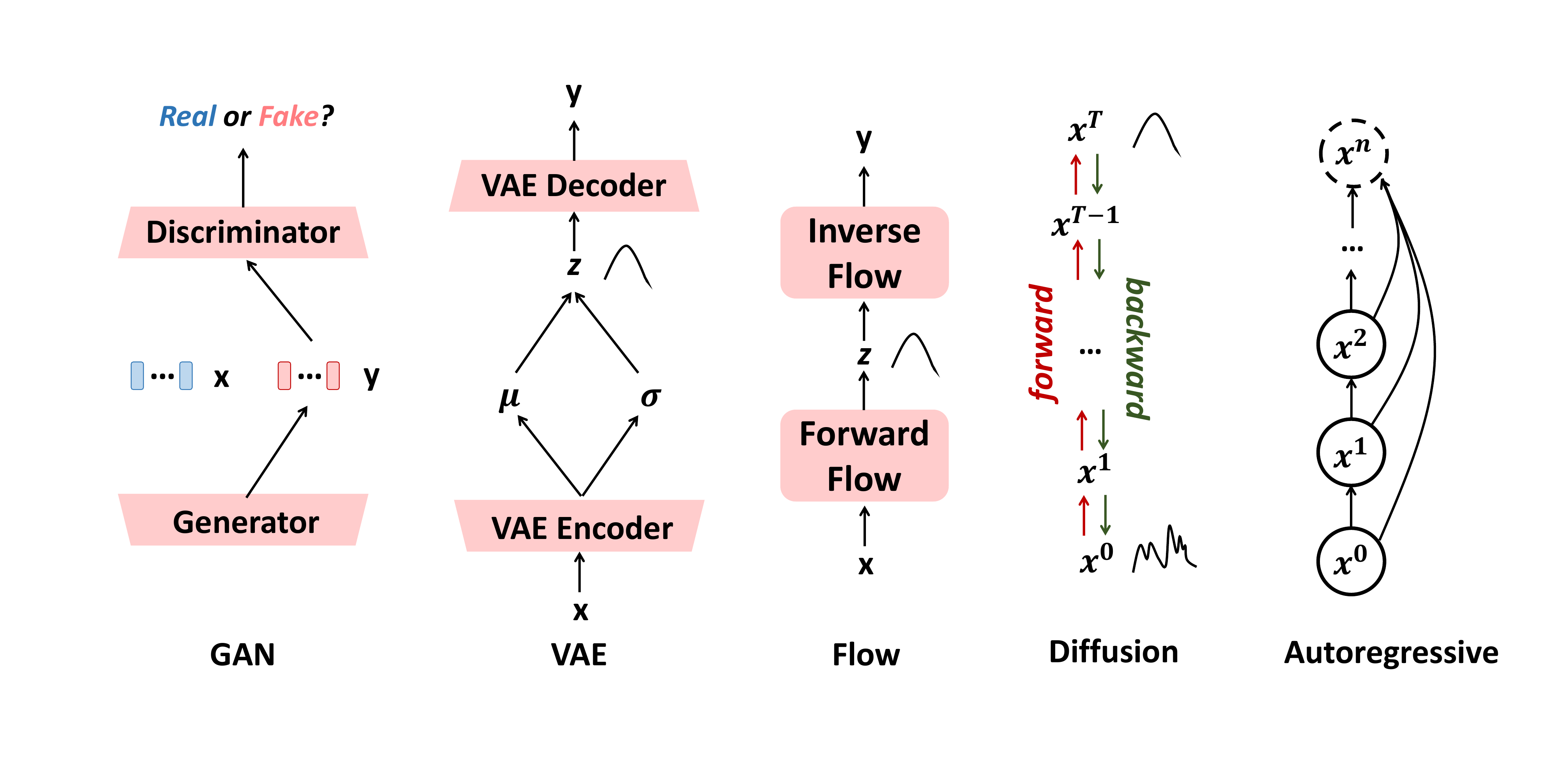}
    \caption{Comparison of multimodal generative model architectures. This figure illustrates the core principles of various generative model architectures: GAN, VAE, Flow, Diffusion, and Autoregressive models. Each architecture has a unique approach to generating data: GAN uses a generator and discriminator for adversarial training; VAE employs an encoder and decoder with a probabilistic approach; Flow models use invertible transformations to map data to latent space and vice versa; Diffusion models gradually add and remove noise to generate samples; Autoregressive models generate data point-by-point conditioned on previously generated points.}

    \label{fig:generative_models}
\end{figure*}

The generative process of models can be described as transforming latent samples \( z \) extracted from a prior distribution \( p_z(z) \) into samples \( x' \) consistent with the target data distribution \( p_{\text{data}}(x) \). Specifically, the latent variables \( z \) are passed through a parameter function, usually implemented as a neural network, learning to map the prior distribution to the target data distribution. The transformed output \( x' \) is then considered a synthetic instance that statistically simulates the characteristics of the original data distribution, potentially corresponding to various modalities such as images, videos, 3D representations, audio, or text.

In the field of multimodal large models (MLMs), generative models play a crucial role in synthesizing new data samples. Major generative methods include Generative Adversarial Networks (GANs)\cite{goodfellow2020generative}, Variational Autoencoders (VAEs)\cite{doersch2016tutorial}, Flow-based Models, Diffusion Models, and Autoregressive Models, as shown in Figure \ref{fig:generative_models}.

\textbf{Generative Adversarial Networks (GANs)}\cite{goodfellow2020generative}: GANs consist of two neural networks: the Generator and the Discriminator. The Generator generates fake samples \( x' \) from latent variables \( z \), trying to deceive the Discriminator, which distinguishes between real samples \( x \) and generated samples. GANs are widely used in image and video generation, as well as high-fidelity audio and text creation.

\textbf{Variational Autoencoders (VAEs)}\cite{doersch2016tutorial}: VAEs include an Encoder and a Decoder. The Encoder maps input data \( x \) to latent space \( z \), learning the mean \( \mu \) and variance \( \sigma \) to generate latent variables. The Decoder reconstructs data \( x' \) from latent space \( z \). VAEs aim to maximize data likelihood while maintaining generative diversity, commonly applied in image synthesis and generating diverse objects.

\textbf{Flow-based Models}\cite{ho2019flow++,han2019clothflow,ping2020waveflow}: Flow models use a series of invertible transformations to map between data space \( x \) and latent space \( z \). Forward flow maps input data \( x \) to latent variables \( z \), while reverse flow reconstructs data \( x' \) from latent variables \( z \). The advantage of flow models lies in precisely modeling data likelihood, often used for high-dimensional data like image and video generation.

\textbf{Diffusion Models\cite{croitoru2023diffusion}}: Diffusion models include a forward process and a reverse process. The forward process gradually transforms data \( x_0 \) into a noise state \( x_T \), while the reverse process denoises \( x_T \) back to data \( x_0 \). Diffusion models learn to reverse the noise process, generating high-quality samples from simple distributions through denoising, particularly suitable for high-resolution image generation and complex multimodal scenarios.

\textbf{Autoregressive Models}: Autoregressive models generate data sequentially, with each step's output depending on the previous step's result. The model generates each data point \( x_t \) conditioned on previous points \( x_{t-1}, x_{t-2}, \ldots, x_1 \). Autoregressive models decompose the joint probability distribution of data into a product of conditional probabilities, widely used in text generation, language modeling, and sequence-based tasks like audio and video generation.

Based on these basic architectures, many significant advancements have emerged recently. Text-to-image generation based on generative models primarily follows two paradigms: diffusion models and VIT-based models\cite{kingma2021variational,yang2023diffusion}. Due to the ease of training, diffusion models have become the mainstream paradigm. Within the diffusion framework, there are pixel-level and latent-level video diffusion models. Diffusion models generate images by using UNet to predict noise, although the process requires iterating multiple time steps (typically denoted as T), becoming very time-consuming as T increases. Additionally, diffusion models cannot control image generation, only generating randomly. To address these issues, Latent Diffusion Model (LDM) proposes a two-stage image generation model: the first stage trains an image encoder-decoder, and the second stage generates images\cite{rombach2022high}. Specifically, LDM simplifies computations by downscaling images to lower scales and adding conditional control modules, injecting image and text features into UNet to guide image generation. Google's Imagen\cite{saharia2022photorealistic} further demonstrates the advantages of pre-trained large models in text-to-image tasks. The model improves noise generation with dynamic sampling and introduces a lightweight UNet model. Cascaded Diffusion Models\cite{ho2022cascaded} improve image clarity and quality by first generating low-resolution images and then progressively upsampling to high resolution. RePaint\cite{lugmayr2022repaint}  proposes a method for image inpainting without training. DALLE2\cite{ramesh2022hierarchical} uses the CLIP model's reverse operation (unCLIP) for image generation, consisting of an image decoder and a prior model, including autoregressive and diffusion-based methods. SDXL\cite{podell2023sdxl} further optimizes diffusion models, improving high-resolution image generation through cascading Base and Refiner models.

Current LLM-based video editing follows schemes similar to Instruct Pix2Pix\cite{brooks2023instructpix2pix}, using LLMs to construct training data more efficiently. Vid2Vid\cite{zhuo2022fast,mallya2020world,wang2019few} is one work involving training data construction by LLMs. This method uses LLM models to generate synthetic video instruction pairs, then trains an editing model to perform controlled video editing using natural language instructions. DiT (Scalable Diffusion Models with Transformers)\cite{peebles2023scalable} is a model widely discussed after Sora, proposing using Transformers instead of UNet structures to enhance generation. 

\subsection{Multimodal Optimization Techniques}

\subsubsection{Multimodal Instruction Tuning (M-IT)}

Multimodal Instruction Tuning (M-IT) is a technique that fine-tunes models on instructions or task descriptions containing multimodal data, enhancing their ability to understand and execute multimodal tasks. Instruction tuning involves fine-tuning pre-trained language models (LLMs) on datasets organized in an instructional format, improving their generalization to unseen tasks \cite{li2023fine,liu2024visual}. This method has been successfully applied in natural language processing models like ChatGPT, InstructGPT, FLAN\cite{xu2024vision}, and OPT-IML\cite{iyer2022opt}.

Traditional supervised fine-tuning relies on large amounts of task-specific data, while prompting methods reduce dependency on large-scale data through prompt engineering, albeit with limited zero-shot performance. Unlike these methods, instruction tuning emphasizes learning to generalize to unseen tasks and closely relates to multi-task prompting. Specifically, datasets constructed for multimodal instruction tuning include specific tasks, input multimodal information, and the expected model output. Through tuning on these multimodal instructions, models better understand how to utilize multimodal capabilities to meet expectations. When extending instruction tuning to multimodal instruction tuning, data and models need adjustments to account for the characteristics of different modal data and their interactions in joint learning. For example, handling vision-text joint tasks requires the model to understand both textual descriptions and related image information. By designing multimodal task descriptions, integrating images and text as input, models use multimodal alignment techniques to learn multimodal features. The core goal of M-IT is to fine-tune models to generalize and handle unseen tasks in various application scenarios, demonstrating stronger adaptability and generalization.

\subsubsection{Multimodal In-Context Learning (M-ICL)}

Multimodal In-Context Learning (M-ICL) enhances models' understanding and processing of multimodal data by providing multimodal contextual information during training or inference\cite{shukor2023beyond,baldassini2024makes}. In-Context Learning (ICL) is an important and emerging capability of large language models (LLMs)\cite{sun2024generative}. ICL achieves few-shot learning and complex task resolution through analogy learning, differing from traditional supervised learning paradigms that require large amounts of data to learn implicit patterns. In ICL settings, LLMs learn from few examples and optional instructions, generalizing to new problems to solve complex and unseen tasks. ICL is training-free and can flexibly integrate into different frameworks' inference stages.

In the context of multimodal large models (MLMs), ICL extends to more modalities, forming Multimodal In-Context Learning (M-ICL). During inference, M-ICL can be achieved by adding a demonstration set (a set of context samples) to the original samples. Specifically, the difference between M-ICL and M-IT lies in constructing datasets with multimodal input-output information, which is contextually related information rather than the expected model response. Through instructions and provided demonstrations, LLMs understand task goals and output templates, generating expected answers. In scenarios teaching LLMs to use external tools, examples usually contain only text information and are more detailed. These examples consist of sequential steps to complete specific tasks, closely related to Chain of Thought (CoT). Combining these techniques, M-ICL extends models' capabilities to handle multimodal tasks and enhances their generalization and adaptability in various application scenarios.

\subsubsection{Multimodal Chain of Thought (M-COT)}

Large Language Models (LLMs) have demonstrated impressive performance in complex reasoning, particularly by using Chain of Thought (CoT) prompts to generate intermediate reasoning chains to infer answers\cite{fengcomprehensive,yin2023survey}. However, existing CoT research primarily focuses on the language modality. Multimodal Chain of Thought (M-COT) is a method that enables models to perform complex reasoning and decision-making through step-by-step derivation and coherent thinking. As noted in previous work, CoT is "a series of intermediate reasoning steps," proven effective in complex reasoning tasks. The core idea of CoT is to prompt the LLM to not only output the final answer but also the reasoning process leading to the answer, akin to human cognitive processes. Inspired by successful experiences in the natural language processing (NLP) field, several research works have extended single-modal CoT to Multimodal CoT (M-CoT).

Zhang et al.\cite{zhang2023multimodal} applied CoT reasoning in multimodal models for the first time. M-COT is a two-stage framework that fine-tunes language models to integrate visual and language representations for better performing multimodal CoT reasoning. In the first stage, the model is fine-tuned with combined visual and language inputs to understand and process multimodal data. In the second stage, the model uses these multimodal representations to incrementally generate intermediate reasoning steps, making coherent and rational decisions in complex tasks. Through this method, M-COT not only enhances the model's reasoning ability in multimodal tasks but also expands its application range in complex scenarios, enabling it to handle tasks that integrate image and text information more effectively.

\section{Overview of Multimodal Models}

In this chapter, we will introduce the generation and basic models that are more influential in the multimodal field. Since most of the models in multimodal generation and underlying models are closed sources, we will not going to make too many statements here. 

\begin{table*}[h!]
\centering
\begin{tabular}{lccccc}
\toprule
\textbf{Model} & \textbf{Application} & \textbf{Input} & \textbf{Output} & \textbf{Designer} & \textbf{Year} \\ 
\midrule
DALL-E & Image Generation & T & I & OpenAI & 2021 \\ 
DALL-E 2 & Image Generation & T & I & OpenAI & 2022 \\ 
DALL-E 3 & Image Generation & T & I & OpenAI & 2023 \\ 
Midjourney & Image Generation & T & I & Midjourney & 2022 \\ 
Imagen & Image Generation & T & I & Google & 2022 \\ 
Imagen 2 & Image Generation & T/I & I & Google & 2023 \\ 
Gen2 & Video Generation and Editing & T/V & V & Runway & 2023 \\ 
Pika & Video Generation & T & V & Pika & 2023 \\ 
Stable Diffusion & Image Generation & T & I & Stability AI & 2022 \\ 
MiniGPT4 & Multimodal Model & T/I & T & KAUST & 2023 \\ 
mPLUG-Owl & Multimodal Model & T/I & T & Alibaba DAMO Academy & 2023 \\ 
LlaVA & Multimodal Model & T/I & T & - & 2023 \\ 
Vmamba & Image Understanding & I & T & UCAS & 2024 \\ 
Open-Sora & Video Generation & T & V & PKU-YuanGroup & 2023 \\ 
Sora & Video Generation & T & V & OpenAI & 2023 \\ 
VideoCrafter & Video Generation and Editing & T/I & V & AILab-CVC & 2023 \\ 
VideoCrafter2 & Video Generation and Editing & T/I & V & AILab-CVC & 2024 \\ 
AudioLM & Audio Generation & T/V & A & Google & 2022 \\ 
SpeechGPT & Speech Understanding and Generation & V/T & T/V & Fudan University & 2023 \\ 
Flamingo & Visual Q and A & I/T & T & DeepMind & 2022 \\ 
InstructBLIP & Visual Q and A & I/T & T & - & 2023 \\ 
Blip-2 & Visual Q and A & I/T & T & - & 2022 \\ 
VideoLlama & Video Understanding & V & T & Alibaba DAMO Academy & 2023 \\ 
VideoLlama 2 & Video and Audio Understanding & V/A & T & Alibaba DAMO Academy & 2024 \\ 
MiniCPM-V & Multimodal Model & I/T & T & Wudaoken Smart, Tsinghua University & 2023 \\ 
MiniCPM-V 2.0 & Multimodal Model & I/T & T & Wudaoken Smart, Tsinghua University & 2024 \\ 
MobileVLM & Multimodal Model & I/T & T & Meituan, Zhejiang University & 2023 \\ 
MobileVLMv2 & Multimodal Model & I/T & T & Meituan, Zhejiang University & 2024 \\ 
Gemini Pro Vision & Multimodal Model & I/T & T & Google & 2023 \\ 
moondream & Multimodal Model & I/T & T & moondream & 2023 \\ 
MM1 & Multimodal Model & I/T & T & Apple & 2024 \\ 
\bottomrule
\\

\end{tabular}
\caption{Comparison of Multimodal Models. Modalities: T = Text, I = Image, V = Video, A = Audio.}
\label{tab:multimodal_models}
\end{table*}

\subsection{Multimodal Generative Models}

Multimodal models have demonstrated significant potential and application prospects in processing and understanding data from different modalities. By analyzing existing multimodal models, it is evident that they have made remarkable progress in generating images, videos, audio, and 3D models. These models achieve cross-modal generation and transformation by handling data from various modalities such as text, images, videos, or audio.

For example, the DALL-E series models (including DALL-E\cite{ramesh2021zero}, DALL-E 2\cite{ramesh2022hierarchical}, DALL-E 3\cite{aghajanyan2023scaling}) developed by OpenAI since 2021 have evolved to generate high-quality images based on textual descriptions. DALL-E models leverage the power of large-scale datasets and transformer architectures to understand and generate highly detailed and creative images from textual inputs. Each iteration has improved on the previous, with DALL-E 3 incorporating more advanced techniques to produce even more coherent and higher-quality images.

Midjourney\cite{oppenlaender2022creativity}, launched in 2022, focuses on high-quality artistic image generation and is widely used in creative design. This model emphasizes generating visually appealing and artistically valuable images, making it popular among artists and designers.

Google's Imagen and Imagen 2\cite{saharia2022photorealistic}, introduced in 2022 and 2023, respectively, have further enhanced image generation capabilities by including editing features. Imagen models allow users to make modifications to generated images, making the process of creating and refining images more interactive and user-friendly.

Stability AI's Stable Diffusion\cite{rombach2022high}, open-sourced in 2022, has propelled community research and application in image generation. The open-source nature of Stable Diffusion has encouraged a wide range of research and development activities, contributing to the rapid advancement of image generation technologies.

The field of video generation has also seen significant breakthroughs. OpenAI and PKU-YuanGroup released Sora and Open-Sora in 2023, capable of generating highly realistic videos. These models extend the capabilities of image generation to the temporal domain, allowing for the creation of dynamic and lifelike video content.

AILab-CVC's VideoCrafter and VideoCrafter2\cite{chen2024videocrafter2}, released in 2023 and 2024 respectively, further optimized video generation and editing functionalities. These models integrate advanced video editing capabilities, making it easier to create and modify video content. Google's AudioLM\cite{borsos2023audiolm}, launched in 2022, excels in audio generation, producing natural sounds, music, and human speech. This model showcases the potential of multimodal generative models in creating high-quality audio content from textual descriptions. Fudan University's SpeechGPT\cite{zhan2024anygpt}, introduced in 2023, combines speech recognition and generation technologies, enhancing the naturalness and accuracy of voice interactions. SpeechGPT integrates advanced language models with speech technologies to provide more fluent and accurate voice interactions.

\subsection{Multimodal Foundation Models}

Multimodal foundation models focus on understanding and aligning multimodal data, enabling more complex reasoning and task execution. DeepMind's Flamingo, launched in 2022, understands image content and answers related questions, excelling in visual question answering tasks. InstructBLIP and Blip-2, introduced between 2022 and 2023, improved visual question answering performance through efficient visual-language pre-training.

DeepMind's Flamingo\cite{alayrac2022flamingo}, launched in 2022, understands image content and answers related questions, excelling in visual question answering tasks. This model is designed to bridge the gap between visual inputs and textual queries, providing accurate and contextually relevant answers.

InstructBLIP and Blip-2\cite{dai2024instructblip,li2023blip}, introduced between 2022 and 2023, improved visual question answering performance through efficient visual-language pre-training. These models leverage large-scale pre-training to enhance their ability to understand and respond to visual queries, making them more effective in various visual question answering scenarios.

Alibaba DAMO Academy's video understanding models, VideoLlama and VideoLlama 2\cite{zhang2023video,cheng2024videollama}, released in 2023 and 2024 respectively, can understand video content and perform tasks such as video subtitle generation and video-audio understanding. These models are designed to process and comprehend complex video data, enabling more sophisticated video analysis and understanding tasks.

Multimodal models like MiniGPT4\cite{zhu2023minigpt}, mPLUG-Owl\cite{ye2023mplug}, and LlaVA\cite{xu2024llava}, developed by KAUST and Alibaba DAMO Academy in 2023, support input and output of text and images. They achieve comprehensive text and image processing through efficient multimodal alignment techniques. These models are designed to handle a variety of multimodal tasks, providing robust performance in text-image alignment and processing.

Meituan and Zhejiang University developed MobileVLM and MobileVLMv2\cite{chu2023mobilevlm,chu2024mobilevlm} in 2023 and 2024, optimizing model performance and application scenarios for mobile platforms, further enhancing multimodal task execution capabilities. These models are specifically designed to be efficient and effective on mobile devices, enabling advanced multimodal functionalities on resource-constrained platforms.

\section{From Multimodal Models to World Models}
Based on current technology, there are two main approaches to constructing a world model from a multimodal model. The first approach relies on rule-based methods and requires only a small amount of data. The second approach, exemplified by OpenAI, involves the use of large datasets. In the following sections, we will introduce these two approaches and explore their potential applicability.

\subsection{3D Generation and Rule Constraints}

3D generation is an essential area in multimodal generation, leading towards world simulators through models that generate realistic 3D models and incorporate rule constraints in the generation process to create highly realistic and controllable virtual environments, akin to the metaverse.

3D generation techniques mainly include explicit representation, implicit representation, and hybrid representation. Explicit representation includes point clouds and meshes, which generate 3D models by precisely describing the geometry of objects. Implicit representations, such as Neural Radiance Fields (NeRF)\cite{mildenhall2021nerf} and implicit surfaces, generate high-quality 3D content by learning latent representations of data. Hybrid representations combine explicit and implicit features, retaining geometric details while offering flexible representation capabilities.

Specific generation methods include Generative Adversarial Networks (GANs), diffusion models, autoregressive models, Variational Autoencoders (VAEs), and normalizing flows. These methods generate realistic 3D data through various mechanisms. For example, GANs generate high-quality 3D models through adversarial training between generators and discriminators; diffusion models generate new samples by simulating the diffusion process of data; autoregressive models generate 3D objects by progressively predicting the conditional probability of each element; VAEs generate data by learning latent representations of input data; normalizing flows use a series of reversible transformations to map simple distributions to data distributions for data generation\cite{shi2023mvdream}.

Optimization-based generation methods use optimization techniques to generate 3D models at runtime, often combining pre-trained networks to optimize 3D models based on user-specified prompts (such as text or images). For example, text-to-3D techniques use textual prompts to guide 3D content generation; image-to-3D techniques reconstruct 3D models from specified images, preserving image appearance and optimizing 3D content geometry. Procedural generation uses predefined rules, parameters, and mathematical functions to create 3D models and textures, including fractal geometry, L-systems, noise functions, and cellular automata\cite{wang2023score,raj2023dreambooth3d}.

Generative novel view synthesis uses generative techniques to predict new views from a single input image, generating new content based on conditional 3D information. Transformer-based methods use multi-head attention mechanisms to gather information from different positions for novel view synthesis; GAN-based methods use 3D point clouds as representations, synthesizing missing regions and generating output images through GANs. These methods have their advantages and application scenarios, and researchers can choose suitable 3D generation techniques based on specific needs\cite{seo2023let,chen2023single}.

Despite significant improvements in 3D generation quality and diversity, current challenges include evaluation, dataset size and quality, representation flexibility, and controllability. Multimodal large models require deeper networks and larger datasets for pre-training. Multimodal large models are often pre-trained on vision and language modalities, and future expansions can include more modalities such as images, text, audio, time, thermal images, etc. Large-scale pre-trained models based on multiple modalities have broader application potential.

\subsection{More Modal Information Leading to Embodied Intelligence}

Another approach to achieving world simulators is through embodied intelligence models. Current multimodal models cover daily information media such as images, text, and audio. However, developing embodied intelligent robots requires expanding these modal information to include coordinate systems, point clouds, and depth, which are crucial for robots to understand and operate in the real world. Integrating these additional modal information into multimodal models can achieve preliminary embodied intelligent robots\cite{gupta2021embodied,howard2019evolving}.

Embodied intelligence involves robots perceiving, understanding, and acting in the physical world. To achieve this, robots need to process and understand data from multiple sensors, such as cameras, LiDAR, and depth sensors\cite{roy2021machine,floreano2004evolution}. These sensors provide information, including coordinate systems, point clouds, and depth maps, enabling robots to construct and understand detailed 3D representations of their surroundings. With this modal information, robots can navigate, recognize objects, and perform tasks in the real world.

By deploying robots in real life, sufficient multimodal information can be collected to achieve comprehensive data collection of the real world. The performance of embodied intelligence can be further enhanced by combining sensor information from multiple sensors on embodied intelligence and edge devices collecting sensor information. With sufficient comprehensive modal information, not only can the performance of embodied intelligence be further improved, but it can also be used to design more realistic world simulators. Such a world simulator can provide robots with a virtual training environment, allowing them to learn and optimize under safe and controlled conditions. Through continuous iteration and optimization, highly intelligent and autonomous embodied intelligent robots can eventually be realized.

\subsection{Incorporating More External Rule Systems}

In the process of constructing world simulators, incorporating more external rule systems is a crucial approach. Humans rely on mathematical, physical, chemical, and biological tools in the objective world, using a series of theorems to derive and predict the outcomes of events that have not yet occurred. For example, when we kick a ball, it will fly in an arc. These predictions based on physical laws help us understand and operate the real world.

Similarly, rule systems can help models achieve state memory and feedback. Suppose a flood breaks the dam; the model needs to infer the subsequent flood state based on rules. These rules stem from human common sense and theorem libraries, summarized from long-term practice and experience. By injecting these conclusions into the model, the model can infer reasonable results with fewer data.

In constructing multimodal large models, integrating external rule systems can significantly enhance model understanding and reasoning capabilities. For example, using mathematical theorems, the model can accurately calculate the trajectory of objects; using physical laws, the model can predict complex environmental changes; using biological knowledge, the model can simulate dynamic changes in ecosystems. These rule systems provide the model with a framework, allowing it to more accurately simulate the real world.

In practical applications, embodied intelligent robots can benefit from these rule systems. When robots collect vast amounts of multimodal data in real life, these data will combine with injected rule systems to enhance robots' prediction and decision-making abilities. For example, when a robot detects rising water levels, it can predict potential flood ranges and impacts based on physical and geographical knowledge and take corresponding actions.

By incorporating these external rule systems, multimodal large models can excel in various application scenarios and achieve more complex and detailed tasks. This approach not only enhances model intelligence but also provides a more solid foundation for future development.

\section{DISCUSSION}

Currently, the development of multimodal large models (MLMs) is still in its early stages, with many challenges and research questions in both related technologies and specific applications.

The perception capabilities of existing MLMs are limited, leading to incomplete or incorrect visual information, further causing subsequent reasoning errors. This situation may result from the compromise between information capacity and computational burden in current models. For example, lowering image resolution and simplifying feature extraction may lead to information loss, affecting the model's overall performance.

The reasoning chain of MLMs is fragile, especially when handling complex multimodal reasoning problems. Even simple tasks sometimes result in incorrect answers due to broken reasoning chains. This indicates that there is still room for improvement in models' understanding and linking of different modal information, requiring more stable and coherent reasoning mechanisms to enhance accuracy and reliability.

The instruction compliance of MLMs needs further improvement. Even after instruction fine-tuning, some MLMs still fail to output expected answers for relatively simple instructions. This suggests that current fine-tuning methods and datasets have not fully covered the various instruction scenarios required by models, necessitating further optimization and expansion of training data.

The issue of object hallucination is prevalent, where MLM outputs responses that do not match the image content, fabricating objects. This not only affects MLM reliability but also reveals deficiencies in visual understanding and semantic generation. Solving this issue requires more precise visual-semantic alignment and verification mechanisms.

Efficient parameter training is another urgent issue. Due to the large capacity of MLMs, efficient parameter training methods can unlock more MLM capabilities under limited computational resources. For example, introducing more effective training strategies and hardware acceleration can significantly reduce model training time and resource consumption, enhancing model application potential.

Currently, there is no truly unified multimodal large model. Although GPT-4o might become the first, significant progress is yet to be seen. This indicates that many technical challenges need to be solved before achieving a truly unified multimodal world simulator. Whether through extensive data training by OpenAI or hierarchical planning with limited data proposed by Meta, or introducing more rules and knowledge bases as mentioned in this paper, these are feasible routes to world models. Fundamentally, extensive data simulate the information humans have encountered since the beginning of civilization, while introducing rules with limited data simulates the rapid learning of descendants using ancestors' summarized experiences and theorems. Both approaches are intuitively reasonable. However, the core issue to be solved currently lies at the micro-level, especially in simplifying attention mechanisms and adapting GPUs to linear attention mechanisms, which can significantly enhance model training efficiency. By deploying edge devices and embodied intelligence to collect data quickly, the arrival of world models is not far off.

\section{SUMMARY}

This work provides a comprehensive overview of the development and challenges of Multimodal Large Models (MLMs), highlighting their potential in advancing artificial general intelligence and world models. It meticulously covers key techniques such as Multimodal Chain of Thought (M-COT), Multimodal Instruction Tuning (M-IT), and Multimodal In-Context Learning (M-ICL), as well as the integration of 3D generation and embodied intelligence. The review emphasizes the importance of external rule systems in enhancing reasoning and decision-making capabilities within MLMs.

This work contributes to the community by offering a detailed analysis of current MLM technologies and their applications, pinpointing the gaps and future research directions necessary for developing a unified multimodal world model.

\bibliographystyle{IEEEtran}
\bibliography{main}

\vfill

\end{document}